\documentclass{article}

\usepackage[utf8]{inputenc} 
\usepackage[T1]{fontenc}    
\usepackage{hyperref}       
\usepackage{url}            
\usepackage{booktabs}       
\usepackage{amsfonts}       
\usepackage{nicefrac}       
\usepackage{microtype}      
\usepackage{amsmath}
\usepackage{subcaption}
\usepackage{graphicx}
\usepackage{multirow}
\usepackage{natbib}
\usepackage{footnote}
\setcitestyle{numbers}
\setcitestyle{square}
\setcitestyle{comma}

\usepackage{color} 
\usepackage{amsmath}
\usepackage{amsfonts}
\usepackage{amssymb}
\usepackage{url}
\usepackage{enumitem}
\usepackage{graphics}
\usepackage{algorithm}
\usepackage{algorithmic}
\usepackage{graphicx}
\usepackage{sidecap}
\usepackage{comment}
\usepackage{wrapfig}
\usepackage{amssymb}
\usepackage{hyperref}
\usepackage{amsfonts,amsthm}
\usepackage{graphicx,amsthm} 
\usepackage{wrapfig}
\usepackage{sidecap}
\usepackage{authblk}
\usepackage{gensymb} 
\allowdisplaybreaks
\usepackage{multicol}
\usepackage{url}            

\usepackage{algorithm}
\usepackage{algorithmic}
\usepackage{bm}
\usepackage{amsmath}
\usepackage{amssymb}
\usepackage{amsfonts}
\usepackage{latexsym}
\usepackage{multirow,tikz,amsmath,comment}
\usepackage{wrapfig}

\usepackage[normalem]{ulem}

\usetikzlibrary{arrows,fit,positioning,shapes,snakes}

\let\oldbibliography\thebibliography
\renewcommand{\thebibliography}[1]{%
  \oldbibliography{#1}%
  \setlength{\itemsep}{1pt}%
}

\newdimen\arrowsize
\pgfarrowsdeclare{arcsq}{arcsq}
{
  \arrowsize=0.2pt
  \advance\arrowsize by .5\pgflinewidth
  \pgfarrowsleftextend{-4\arrowsize-.5\pgflinewidth}
  \pgfarrowsrightextend{.5\pgflinewidth}
}
{
  \arrowsize=1.5pt
  \advance\arrowsize by .5\pgflinewidth
  \pgfsetdash{}{0pt} 
  \pgfsetroundjoin   
  \pgfsetroundcap    
  \pgfpathmoveto{\pgfpoint{0\arrowsize}{0\arrowsize}}
  \pgfpatharc{-90}{-140}{4\arrowsize}
  \pgfusepathqstroke
  \pgfpathmoveto{\pgfpointorigin}
  \pgfpatharc{90}{140}{4\arrowsize}
  \pgfusepathqstroke
}

\newtheorem{Proposition}{Proposition}

\title{Causal Generative Domain Adaptation Networks}

\author{Mingming Gong\thanks{Equal Contribution}~\thanks{Department of Biomedical Informatics, University of Pittsburgh, USA}~$^\ddagger$~~~~Kun Zhang$^*$\thanks{Department of Philosophy, Carnegie Mellon University, USA}~~~~Biwei Huang$^\ddagger$ ~~~~Clark Glymour$^\ddagger$ ~~~~ Dacheng Tao\thanks{UBTECH Sydney AI Center, The University of Sydney, Australia}~~~~ Kayhan Batmanghelich$^\dagger$}

\begin{document}

\maketitle

\begin{abstract}

An essential problem in domain adaptation is to understand and make use of distribution changes across domains. For this purpose, we first propose a flexible Generative Domain Adaptation Network (G-DAN) with specific latent variables to capture changes in the generating process of features across domains. By explicitly modeling the changes, one can even generate data in new domains using the generating process with new values for the latent variables in G-DAN. In practice, the process to generate all features together may involve high-dimensional latent variables, requiring dealing with distributions in high dimensions and making it difficult to learn domain changes from few source domains. Interestingly, by further making use of the causal representation of joint distributions, we then decompose the joint distribution into separate modules, each of which involves different low-dimensional latent variables and can be learned separately, leading to a Causal G-DAN (CG-DAN). This improves both statistical and computational efficiency of the learning procedure. Finally, by matching the feature distribution in the target domain, we can recover the target-domain joint distribution and derive the learning machine for the target domain. We demonstrate the efficacy of both G-DAN and CG-DAN in domain generation and cross-domain prediction on both synthetic and real data experiments. 



\end{abstract}

\section{Introduction}

In recent years supervised learning has achieved great success in various real-world problems, such as visual recognition, speech recognition, and natural language processing. However, the predictive model learned on training data may not generalize well when the distribution of test data is different. For example, a predictive model trained with data from one hospital may fail to produce reliable prediction in a different hospital due to distribution change. Domain adaption (DA) aims at learning models that can predict well in test (or target) domains by transferring proper information from source to target domains \cite{pan2010survey,Jiang08}. In this paper, we are concerned with a difficult scenario called unsupervised domain adaptation, where no labeled data are provided in the target domain.

Let $X$ denote features and $Y$ the class label. If the joint distribution $P_{XY}$ changes arbitrarily, apparently the source domain may not contain useful knowledge to help prediction in the target domain. Thus, various constraints on how distribution changes have been assumed for successful transfer. For instance, a large body of previous methods assume that the marginal distribution $P_X$ changes but $P_{Y|X}$ stays the same, i.e., the \textit{covariate shift} situation ~\cite{Shimodaira00,Huang07,Sugiyama08,cortes2008sample}. In this situation, correcting the shift in $P_X$ can be achieved by reweighting source domain instances \cite{Huang07,Sugiyama08} or learning a domain-invariant representation \cite{Pan11,6751205,ganin2016domain}. 

This paper is concerned with a more difficult situation, namely \textit{conditional shift} \cite{Zhang13_targetshift}, where $P_{X|Y}$ changes, leading to simultaneous changes in $P_X$ and $P_{Y|X}$. In this case, $P_{XY}$ in the target domain is generally not recoverable because there is no label information in the target domain. Fortunately, with appropriate constraints on how $P_{X|Y}$ changes, we may recover $P_{XY}$ in the target domain by matching only $P_X$ in the target domain. For example, location-scale transforms in $P_{X|Y}$ have been assumed and the corresponding identifiability  conditions have been established \cite{Zhang13_targetshift,GonZhaLiuTaoSch16}. Despite its success in certain applications, the location-scale transform assumption may be too restrictive in many practical situations, calling for more general treatment of distribution changes.

How can we model and estimate the changes in $P_{X|Y}$ without assuming strong parametric constraints? To this end, we first propose a Generative Domain Adaptation Network (G-DAN) with specific latent variables $\boldsymbol{\theta}$ to capture changes in $P_{X|Y}$. Specifically, the proposed network implicitly represents $P_{X|Y}$ by a functional model $X=g(Y,E,\boldsymbol{\theta})$, where the latent variables $\boldsymbol{\theta}$ may take different values across domains, and the independent noise $E$ and the function $g$ are shared by all domains. This provides a compact and nonparametric representation of changes in $P_{X|Y}$, making it easy to capture changes and find new sensible values of $\boldsymbol{\theta}$ to generate new domain data. Assuming $\boldsymbol{\theta}$ is low-dimensional, we provide the necessary conditions to recover $\boldsymbol{\theta}$ up to some transformations from the source domains. In particular, if the changes can be captured by a single latent variable, we show that under mild conditions the changes can be recovered from a single source domain and an unlabeled target domain. We can then estimate the joint distribution in the target domain by matching only the feature distribution, which enables target-domain-specific prediction. Furthermore, by interpolation in the $\boldsymbol{\theta}$ space and realization of the noise term $E$, one can straightforwardly generate sensible data in new domains. 

However, as the number of features increases, modeling $P_{X|Y}$ for all features $X$ jointly becomes much more difficult . To circumvent this issue, we propose to factorize joint distributions of features and the label, according to the causal structure underlying them. Each term in the factorization aims at modeling the conditional distribution of one or more variables given their direct causes, and accordingly, the latent variables $\boldsymbol{\theta}$ are decomposed into disjoint, unrelated subsets \cite{Pearl00}. Thanks to the modularity of causal systems, the latent variables for those terms can be estimated separately, enjoying a ``divide-and-conquer'' advantage. With the resulting Causal G-DAN (CG-DAN), it is then easier to interpret and estimate the latent variables and find their valid regions to generate new, sensible data. Our contribution is mainly two fold:
\begin{itemize}[noitemsep,nolistsep,topsep=-3pt]
 \item Explicitly capturing invariance and changes across domains by exploiting latent variable $\boldsymbol{\theta}$ in a meaningful way.
 \item Further facilitating the learning of $\boldsymbol{\theta}$ and the generating process in light of causal representations.
\end{itemize}

\section{Related Work}
Due to the distribution shift phenomenon, DA has attracted a lot of attention in the past decade, and here we focus on related unsupervised DA methods. In the covariate shift scenario, where $P_X$ changes but $P_{Y|X}$ stays the same, to correct the shift in $P_X$, traditional approaches reweight the source domain data by density ratios of the features \cite{Shimodaira00,Sugiyama08,Huang07,Cortes10,yu2012analysis}. However, such methods require the target domain to be contained in the support of the source domain, which may be restrictive in many applications.  Another collection of methods searches for a domain-invariant representation that has invariant distributions in both domains. These methods rely on various distance discrepancy measures as objective functions to match representation distributions; typical ones include maximum mean discrepancy (MMD) \cite{Pan11,6751205}, separability measured by classifiers \cite{ganin2016domain}, and optimal transport \cite{courty2017joint}. Moreover, the representation learning architecture has developed from shallow architectures such as linear projections \cite{6751205} to deep neural networks \cite{icml2015_long15,ganin2016domain}. 

Recently, another line of researches has attempted to address a more challenging situation where $P_X$ and $P_{Y|X}$ both change across domains. A class of methods to deal with this situation assumes a (causal) generative model $Y\rightarrow X$, factorizes the joint distribution following the causal direction as $P_{XY} = P_YP_{X|Y}$, and considers the changes in $P_Y$ and $P_{X|Y}$ separately. Assuming that only $P_Y$ changes, one can estimate $P_Y$ in the target domain by matching the feature distribution \cite{Storkey09,Zhang13_targetshift,Iyer14}. Further works also consider changes in $P_{X|Y}$, known as \textit{generalized target shift}, and proposed representation learning methods with identifiability justifications, i.e., the learned representation $\tau(X)$ has invariant $P_{\tau(X)|Y}$ across domains if $P_{\tau(X)}$ is invariant after correction for $P_Y$ \cite{Zhang13_targetshift,GonZhaLiuTaoSch16}. This also partially explains why previous representation learning methods for correcting covariate shift work well in the \textit{conditional shift} situation where only $P_{X|Y}$ changes but $P_Y$ remains the same. To better match joint distributions across domains, recent methods focus on exploring more powerful distribution discrepancy measures and more expressive representation learning architectures \cite{courty2017joint,long2017deep}. 

In the causal discovery (i.e., estimating the underlying causal structure from observational data) field~\cite{Spirtes00,Pearl00}, some recent methods tried to learn causal graphs by representing functional causal models using neural networks \cite{goudet2017learning}. Such methods leveraged conditional independences and different complexities of the distribution factorizations in the data from a fixed distribution to learn causal structures. In contrast, the purpose of our work is not  causal discovery, but to make use of generative models for capturing domain-invariant and changing information and  benefit from the modularity property of causal models \cite{Pearl00} to understand and deal with  changes in distributions.

\section{Generative Domain Adaptation Network (G-DAN)} \label{Sec:general}

In unsupervised DA, we are given $m$ source domains $\mathcal{D}_s=\{(x_i^s,y_i^s)\}_{i=1}^{n_s}\sim P^s_{XY}$, where $s\in\{1,\cdots,m\}$ and a target domain $\mathcal{D}_t=\{x_i^t\}_{i=1}^{n_t}$ of $n_t$ unlabeled examples sampled from $P^t_X$. The goal is to derive a learning machine to predict target-domain labels by making use of information from all domains. To achieve this goal, understanding how the generating process of all features $X$ given labels $Y$ changes across domains is essential. Here we have assumed that $Y$ is a cause of $X$, as is the case in many classification problems~\cite{Zhang13_targetshift}. 


\begin{wrapfigure}{r}{0.35\textwidth}
\begin{center}
\begin{tikzpicture}[boxx1/.style={draw,rounded corners,minimum height=2cm,text width=2cm,align=center,text centered}]
\node[boxx1] (a1) at(0,0.3) {$g$ (represented by NN)};
\node[inner sep = 1pt, circle,draw, left=of a1.150,font=\bfseries] (aux11) {$Y$};
\node[left=of a1.175,font=\bfseries] (aux31) {$E$};
\node[left=of a1.210,font=\bfseries] (aux41) {$\boldsymbol{\theta}$};
\node[inner sep = 1pt, circle,draw, right=of a1.5,font=\bfseries] (c) {$X$};
\draw[->] (aux11) -- (a1.150);
\draw[->] (aux31) -- (a1.175);
\draw[->] (aux41) -- (a1.200);
\draw[->] (a1.5) -- (c.west);
\end{tikzpicture}
\caption{G-DAN.} \label{fig:NN}
\vspace{-10pt}
\end{center}
\end{wrapfigure}
In this section, we propose a Generative Domain Adaptation Network (G-DAN), as shown in Fig. \ref{fig:NN}, to model both invariant and changing parts in $P_{X|Y}$ across domain for the purpose of DA. G-DAN uses specific latent variables $\boldsymbol{\theta}\in\mathbb{R}^d$ to capture the variability in $P_{X|Y}$. In the $s$-th domain, $\boldsymbol{\theta}$ takes value $\theta^{(s)}$ and thus encodes domain-specific information. In Fig. \ref{fig:NN}, the network specifies a distribution $Q_{X|Y;\boldsymbol{\theta}}$ by the following functional model:
\begin{equation}\label{Eq:GDAN}
X=g(Y, E; \boldsymbol{\theta}),
\end{equation}
which transforms random noise $E\sim Q_E$ to $X\in \mathbb{R}^D\sim Q_{X|Y;\boldsymbol{\theta}}$, conditioning on $Y$ and $\boldsymbol{\theta}$. $Q_{X|Y;\boldsymbol{\theta}=\theta^{(s)}}$ is trained to approximate the conditional distribution in the $s$-th domain $P_{X|Y,S=s}$, where $S$ is the domain index. $E$ is independent of $Y$ and has a fixed distribution across domains. $g$ is a function represented by a neural network (NN) and shared by all domains.  Note that the functional model can be seen as a way to specify the conditional distribution of $X$ given $Y$. For instance, consider the particular case where $X$ and $E$ have the same dimension and the $E$ can be recovered from $X$ and $Y$. Then by the formula for change of variables, we have 
$Q_{YX} = Q_{YE}/|\partial (Y,X)/\partial (Y,E)| = Q_{Y}Q_E/|\partial X/\partial E| $; as a consequence, $Q_{X|Y} = Q_{XY}/Q_{Y} = Q_E/|\partial X/\partial E| $.

G-DAN provides a compact way to model the distribution changes across domains. For example, in the generating process of images, illumination $\boldsymbol{\theta}$ may change across different domains, but the mechanism $g$ that maps the class labels $Y$, illumination $\boldsymbol{\theta}$ (as well as other quantities that change across domains), and other factors $E$ into the images $X$ is invariant. This may resemble how humans understand different but related things. We can capture the invariance part, understand the difference part, and even generate new virtual domains according to the perceived process.

\subsection{Identifiability}\label{subsec:identifiability}
Suppose the true distributions $P_{X|Y;\boldsymbol{\eta}}$ across domains were generated by some function with changing (effective) parameters $\boldsymbol{\eta}$--the function is fixed and $\boldsymbol{\eta}$ changes across domains.  It is essential to see whether $\boldsymbol{\eta}$  is identifiable given enough source domains. Here identifiability is about whether the estimated $\hat{\boldsymbol{\theta}}$ can capture all distribution changes, as implied by the following proposition.

\begin{Proposition}\label{Theo:identi11}
Assume that $\boldsymbol{\eta}$ in $P_{X|Y;\boldsymbol{\eta}}$ is identifiable \cite{hoel1954introduction}, i.e., $ P_{X|Y,\boldsymbol{\eta}=\eta_1}=P_{X|Y,\boldsymbol{\eta}=\eta_2}\Rightarrow {\eta}_1={\eta}_2$ for all suitable ${\eta}_1,~{\eta}_2\in \mathbb{R}^d$. Then as $n_s\rightarrow \infty$, if ${P}^s_{X|Y}=Q^s_{X|Y}$, the estimated $\hat{\boldsymbol{\theta}}$ captures the variability in $\boldsymbol{\eta}$ in the sense that $\hat{{\theta}}_1=\hat{{\theta}}_2\Rightarrow {\eta}_1={\eta}_2$.
\end{Proposition}

A complete proof of Proposition \ref{Theo:identi11} can be found in Section S1 of Supplementary Material. The above results says that different values of $\boldsymbol{\eta}$ correspond to different values of $\hat{\theta}$ once $Q_{X|Y;\boldsymbol{\theta}}$ perfectly fits the true distribution. It should be noted that if one further enforces that different $\boldsymbol{\theta}$ correspond to different $Q_{X|Y;\boldsymbol{\theta}}$ in (\ref{Eq:GDAN}), then we have $\hat{{\theta}}_1=\hat{{\theta}}_2\Leftrightarrow {\eta}_1={\eta}_2$. That is, the true $\boldsymbol{\eta}$ can be estimated up to a one-to-one mapping.


If $\boldsymbol{\theta}$ is high-dimensional, intuitively, it means that the changes in conditional distributions are complex; in the extreme case, it could change arbitrarily across domains. To demonstrate that the number of required domains increases with the dimensionality of $\boldsymbol{\theta}$, for illustrative purposes, we consider a restrictive, parametric case where the expectation of $P_{X|Y;\boldsymbol{\theta}}$ depends linearly on $\boldsymbol{\theta}$.


\begin{Proposition}\label{Theo:identi1}
Assume that for all suitable $\boldsymbol{\theta}\in \mathbb{R}^d$ and function $g$, $Q_{X|Y;\boldsymbol{\theta},g}$ satisfies $E_{Q_{X|Y;\boldsymbol{\theta},g}}[X]=A\boldsymbol{\theta}+h(Y)$, where $A\in\mathbb{R}^{D\times d}$. Suppose the source domain data were generated from $Q_{X|Y;\boldsymbol{\theta}^*,g^*}$ implied by model (\ref{Eq:GDAN}) with true $\boldsymbol{\theta}^*$ and $g^*$. The corresponding expectation is $A^*\boldsymbol{\theta}^*+h^*(Y)$. Denote $\Theta^*\in \mathbb{R}^{(d+1)\times m}$ as a matrix whose $s$-th column is $[\theta^{(s)};1]$ and $A^*_{aug}$ as the matrix $[A^*,h^*(Y)]$. Assume that $rank(\Theta^*)= d+1$ (a necessary condition is $m\geq d+1$) and $rank(A^*_{aug})=d+1$, if $P^s_{X|Y}=Q^s_{X|Y}$, the estimated $\hat{\boldsymbol{\theta}}$ is a one-to-one mapping of $\boldsymbol{\theta^*}$.
\end{Proposition}
A proof of Proposition \ref{Theo:identi1} can be found in Section S2 of Supplementary Material. If the number of domains $m$ is smaller than the dimension of $\boldsymbol{\theta}$, the matrix $\Theta^*$ and the estimated $\hat{A}$ will be rank deficient. Therefore, only part of changes in $\boldsymbol{\theta}$ can be recovered. 

In practice, we often encounter a very challenging situation when there is only one source domain. In this case, we must incorporate the target domain to learn the distribution changes. However, due to the absence of labels in the target domain, the identifiability of $\boldsymbol{\theta}$ and the recovery of the target domain joint distribution needs more assumptions. One possibility is to use linear independence constraints on the modeled conditional distributions, as stated below.

\begin{Proposition} \label{Theo:identi_single}
Assume $Y\in\{1,\cdots,C\}$. Denote by $Q_{X|Y;\theta}$ the conditional distribution of $X$ specified by $X=g(Y,E;\boldsymbol{\theta})$. Suppose for any $\boldsymbol{\theta}$ and $\boldsymbol{\theta}'$, the elements in the set $\big\{\lambda Q_{X|Y=c;\boldsymbol{\theta}}+\lambda' Q_{X|Y=c;\boldsymbol{\theta}'}\,;\,c=1, ..., C\big\}$ are linearly independent for all $\lambda$ and $\lambda'$  such that $\lambda^2 + \lambda'^2 \neq 0$. 
If the marginal distributions of the generated $X$ satisfy $Q_{X|\boldsymbol{\theta}}=Q_{X|\boldsymbol{\theta}'}$, then $Q_{X|Y;\boldsymbol{\theta}}=Q_{X|Y;\boldsymbol{\theta}'}$ and the corresponding marginal distributions of $Y$ also satisfies $Q_Y=Q'_Y$.
\end{Proposition}

A proof is given in Section S3 of Supplementary Material. The above conditions enable us to identify $\boldsymbol{\theta}$ as well as recover the joint distribution in the target domain in the single-source DA problem.

\subsection{Model Learning}\label{SubSec:ad_train}
To estimate $g$ and $\boldsymbol{\theta}$ from empirical data, we adopt the adversarial training strategy \cite{goodfellow2014generative,li2015generative}, which minimizes the distance between the empirical distributions in all domains and the empirical distribution of the data sampled from model (\ref{Eq:GDAN}). Because our model involves an unknown $\boldsymbol{\theta}$ whose distribution is unknown, we cannot directly sample data from the model. To enable adversarial training, we reparameterize $\boldsymbol{\theta}$ by a linear transformation of the one-hot representation $\mathbf{1}_s$ of domain index $s$, i.e., $\boldsymbol{\theta}=\Theta \mathbf{1}_s$, where $\Theta$ can be learned by adversarial training.

We aim to match distributions in all source domains and the target domain. 
In particular, in the $s$-th source domain, we estimate the model by matching the joint distributions $P^s_{XY}$ and $Q^s_{XY}$, where $Q^s_{XY}$ is the joint distribution of the generated features $X=g(Y,E,\theta_s)$ and labels $Y$. Specifically, we use the Maximum Mean Discrepancy (MMD)  \cite{Huang07,Song13_embedding,li2015generative} to measure the distance between the true and model distributions:
\begin{flalign}
 J_{kl}^s = \|E_{(X,Y)\sim P_{XY}^s}[\phi(X)\otimes \psi(Y)] - E_{(X,Y)\sim Q_{XY}^s}[\phi(X)\otimes \psi(Y)]\|_{\mathcal{H}_x\otimes \mathcal{H}_y}^2,\label{Eq:joint_mmd}
\end{flalign}
where $\mathcal{H}_x$ denotes a characteristic Reproducing Kernel Hilbert Space (RKHS) on the input feature space $\mathcal{X}$ associated with a kernel $k(\cdot,\cdot): \mathcal{X} \times \mathcal{X} \rightarrow \mathbb{R}$, $\phi$ the associated mapping such that $\phi(x) \in \mathcal{H}_x$, and $H_y$ the RKHS on the label space $\mathcal{Y}$ associated with kernel $l(\cdot,\cdot): \mathcal{Y} \times \mathcal{Y} \rightarrow \mathbb{R}$ and mapping $\psi$.
In practice, given a mini-batch of size $n$ consisting of $\{(x^s_i,y^s_i)\}_{i=1}^{n}$ sampled from the source domain data and $\{(\hat{x}^s_i,\hat{y}^s_i)\}_{i=1}^{n}$ sampled from the generator, we need to estimate the empirical MMD for gradient evaluation and optimization:
\begin{flalign}
\hat{J}^s_{kl}
=&~\frac{1}{n^2}\sum_{i}^{n}\sum_{j}^{n}k(x^s_i,x^s_j)l(y^s_i,y^s_j)
-\frac{2}{n^2}\sum_{i}^{n}\sum_{j}^{n}k(x^s_i,\hat{x}^s_j)l(y_i,\hat{y}^s_j)\nonumber\\
&+\frac{1}{n^2}\sum_{i}^{n}\sum_{j}^{n}k(\hat{x}^s_i,\hat{x}^s_j)l(\hat{y}^s_i,\hat{y}^s_j)\nonumber
\end{flalign}
In the target domain, since no label is present, we propose to match the marginal distributions $P^t_{X}$ and $Q^t_{X}$, where $Q^t_{X}$is the marginal distribution of the generated features $X=g(Y,E,\theta_{t})$, where $\theta_{t}$ is the $\boldsymbol{\theta}$ value in the target domain. The loss function is $M_{k}=\|E_{X\sim P_{X}^t}[\phi(X)] - E_{X\sim Q_{X}^t}[\phi(X)]\|_{\mathcal{H}_x}^2$, and its empirical version $\hat{M}_{k}$ is similar to $\hat{J}^s_{kl}$ except that the terms involving the kernel function $l$ are absent. Finally, the function $g$ and the changing parameters (latent variables) $\Theta$ can be estimated by $\min_{g,\Theta} J_{kl}^s+\alpha M_{k}$, where $\alpha$ is a hyperparameter that balances the losses on source and target domains. In the experiments, we set $\alpha$ to 1.


\section{Causal Generative Domain Adaptation Networks (CG-DAN)}
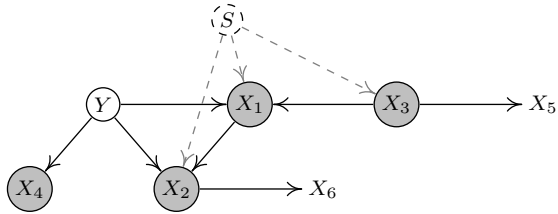
\begin{figure}
\centering
{\hspace{0cm}\begin{tikzpicture}[scale=.75, line width=0.5pt, inner sep=0.2mm, shorten >=.1pt, shorten <=.1pt]
\draw (0, 0) node(2) [circle, draw, fill=black!25]  {{\footnotesize\,${X}_3$\,}};
  \draw (-2.6, 0) node(1)[circle, draw, fill=black!25]  {{\footnotesize\,${X}_1$\,}};
\draw (2.6, 0) node(3)  {{\footnotesize\,${X}_5$\,}};
\draw (-5.2, 0) node(5)[circle, draw]  {{\footnotesize\,$Y$\,}};
\draw (-3.9, -1.5) node(6)[circle, draw, fill=black!25]  {{\footnotesize\,${X}_2$\,}};
\draw (-1.3, -1.5) node(7) {{\footnotesize\,${X}_6$\,}};
\draw (-6.5, -1.5) node(8) [circle, draw, fill=black!25]  {{\footnotesize\,${X}_4$\,}};
\draw (-3, 1.5) node(9) [circle, draw, dashed]  {{\footnotesize\,$S$\,}};
  \draw[-arcsq] (2) -- (1); 
  \draw[-arcsq] (2) -- (3); 
        \draw[-arcsq] (5) -- (1);
            \draw[-arcsq] (5) -- (6);
                        \draw[-arcsq] (1) -- (6);
                        \draw[-arcsq] (6) -- (7);
                        \draw[-arcsq] (5) -- (8);
                        \draw[-arcsq,gray, dashed] (9) -- (1);
                        \draw[-arcsq,gray, dashed] (9) -- (2);
                        \draw[-arcsq,gray, dashed] (9) -- (6);
\end{tikzpicture}} 
\caption{A causal graph over $Y$ and features $X_i$. $Y$ is the variable to be predicted, and nodes in gray are in its Markov Blanket. $S$ denotes the domain index; a direct link from $S$ to a variable indicates that the generating process for that variable changes across domains. Here the generating processes for $Y$, $X_1$, $X_2$, and $X_3$ vary across domains.}
\vspace{-10pt}
\label{fig:MB}
\end{figure}

The proposed G-DAN models the class-conditional distribution, $P_{X|Y}^s$, for all features $X$ jointly. As the dimensionality of $X$ increases, the estimation of joint distributions requires more data in each domain to achieve a certain accuracy. In addition, more domains are needed to learn the distribution changes.

To circumvent this issue, we propose to factorize joint distributions of features and the label according to the causal structure. Each term in the factorization aims at modeling the conditional distribution of one variable (or a variable group) given the direct causes, which involves a smaller number of variables \cite{Pearl00}. Accordingly, the latent variables $\boldsymbol{\theta}$ can be decomposed into disjoint, unrelated subsets, each of which has a lower dimensionality.

\subsection{Causal Factorization}
We model the distribution of $Y$ and relevant features following a causal graphical representation. We assume that the causal structure is fixed, but the generating process, in particular, the function or parameters, can change across domains. According to the modularity properties of a causal system, the changes in the factors of the factorization of the joint distribution are independent of each other \cite{Spirtes00,Pearl00}. We can then enjoy the ``divide-and-conquer'' strategy to deal with those factors separately.

Fig. \ref{fig:MB} shows a Directed Acyclic Graph (DAG) over $Y$ and features $X_i$. According to the Markov factorization \cite{Pearl00}, the distribution over $Y$ and the variables in its Markov Blanket (nodes in gray) can be factorized according to the DAG as $P_{XY} = P_Y P_{X_1|Y,X_3} P_{X_2|X_1,Y} P_{X_3} P_{X_4|Y}$.
For the purpose of DA, it suffices to consider only the Markov blanket $MB(Y)$ of $Y$, since $Y$ is independent of remaining variables given its Markov blanket. By considering only the Markov blanket, we can reduce the complexity of the model distribution without hurting the prediction performance. 

We further adopt functional causal model (FCM) \cite{Pearl00} to specify such conditional distributions. The FCM (a tuple $\langle\mathbf{F},P \rangle$) consists of a set of equations $\mathbf{F}=(F_1,\ldots,F_D)$:
\begin{flalign}
F_i: X_i  =  g_i(\mathbf{PA}_i,E_i,\boldsymbol{\theta}_i),~~i=1,\ldots,D,
\label{eq:sems}
\end{flalign}
and a probability distribution $P$ over $\mathbf{E}=(E_1,\ldots,E_D)$. $\mathbf{PA}_i$ denotes the direct causes of $X_i$, and $E_i$ represents noises due to unobserved factors. Modularity implies that parameters in different FCMs are unrelated, so the latent variable $\boldsymbol{\theta}$ can be decomposed into disjoint subsets $\boldsymbol{\theta}_i$, each of which only captures the changes in the corresponding conditional distribution $P_{X_i|\mathbf{PA}_i}$. $g_i$ encodes the invariance part in $P_{X_i|\mathbf{PA}_i}$ and is shared by all domains. $E_i$ are required to be jointly independent. Without loss of generality, we assume that all the noises $\{E_i\}_{i=1}^n$ follow Gaussian distributions. (A nonlinear function of $E_i$, as part of $g_i$, will change the distribution of the noise.)

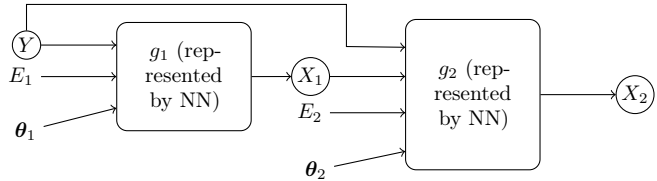
\begin{figure}
 \centering
 \begin{tikzpicture}[scale=0.8, every node/.style={scale=0.8}, boxx1/.style={draw,rounded corners,minimum height=1.8cm,text width=2cm,align=center,text centered}]
 \node[boxx1] (a1) at(0,0.3) {$g_1$ (represented by NN)};
 \node[inner sep = 1pt, circle,draw, left=of a1.155,font=\bfseries] (aux11) {$Y$};
 \node[left=of a1.180,font=\bfseries] (aux31) {$E_1$};
 \node[left=of a1.220,font=\bfseries] (aux41) {$\boldsymbol{\theta}_1$};
 \node[boxx1,minimum height=2.5cm] (a) at(4.8,0) {$g_2$ (represented by NN)};
 \node[left=of a.145,font=\bfseries] (aux1) {};
 \node[inner sep = 1pt, circle,draw, left=of a.165,font=\bfseries] (aux2) {${X}_1$};
 \node[left=of a.195,font=\bfseries] (aux3) {$E_2$};
 \node[left=of a.230,font=\bfseries] (aux4) {$\boldsymbol{\theta}_2$};
 \node[inner sep = 1pt, circle,draw, right=of a,font=\bfseries] (c) {${X}_2$};
 \draw[->] (a1.east) -- (aux2.west);
 \draw[->] (aux11) -- (a1.155);
 \draw[->] (aux31) -- (a1.180);
 \draw[->] (aux41) -- (a1.205);
 \draw(aux11) |- (-.9,1.5) -|  (2.7,.8) [->]--  (a.145);
 \draw[->] (aux2) -- (a.165);
 \draw[->] (aux3) -- (a.195);
 \draw[->] (aux4) -- (a.220);
 \draw[->] (a.east) -- (c.west);
 \end{tikzpicture}
 \caption{CG-DAN for with modules $Y\rightarrow X_1$ and $(Y,X_1) \rightarrow X_2$.} \label{fig:cNN}
\vspace{-0.5cm}
 \end{figure}
\subsection{Learning Given the Network Structure}\label{Subsec:net_cons}

Based on (\ref{eq:sems}), we employ a neural network to model each $g_i$ separately, and construct a constrained generative model according to the causal DAG, which we call a Causal Generative Domain Adaptation Network (CG-DAN). Fig. \ref{fig:cNN} gives an example network constructed on $Y$, $X_1$, and $X_2$ in Fig \ref{fig:MB}. (For simplicity we have ignored $X_3$.)

Because of the causal factorization, we can learn each conditional distribution $Q_{X_i|\mathbf{PA}_i,\boldsymbol{\theta}_i}$ separately by the adversarial learning procedure described in section \ref{SubSec:ad_train}. To this end, let us start by using the Kullback-Leibler (KL) divergence to measure the distribution distance, and the function to be minimized can be written as

\begin{flalign} \nonumber
\setlength{\abovecaptionskip}{0pt}
\setlength{\belowcaptionskip}{5pt}
\vspace{5cm}
\texttt{KL}(P^s_{XY} \,||\, Q^{s}_{XY|\boldsymbol{\theta}}) &= \mathbb{E}_{P^s_{XY}} \Big\{  \log \frac{P^{s}_{Y , X_1 , \cdots, X_D}}{ Q^{s}_{Y , X_1, \cdots, X_D| \boldsymbol{\theta}} }   \Big\}\nonumber\\
&=\mathbb{E}_{P^s_{XY}}\Big\{  \log \frac{P^s_Y\Pi_{i=1}^D P^s_{X_i|\mathbf{PA}_i} }{Q^s_Y\Pi_{i=1}^D Q^s_{X_i|\mathbf{PA}_i,\boldsymbol{\theta}_i}} \Big\} \nonumber\\ \label{Eq:KL1}
 &=\sum_{i=1}^D\texttt{KL}(P^{s}_{ X_i \,|\, \mathbf{PA}_i}\,||\, Q^{s}_{X_i\,|\,\mathbf{PA}_i, \boldsymbol{\theta}_i})\nonumber\\
 &=\sum_{i=1}^D\texttt{KL}(P^{s}_{ X_i \,|\, \mathbf{PA}_i}\,||\, Q^{s}_{X_i\,|\,\mathbf{PA}_i, \boldsymbol{\theta}_i}P^s_Y),\nonumber
\end{flalign}


where each term corresponds to a conditional distribution of relevant variables given all its parents. It can be seen that the objective function is the sum of the modeling quality of each module. For computational convenience, we can use MMD (\ref{Eq:joint_mmd}) to replace the KL divergence.

\paragraph{Remark} It is worthwhile to emphasize the advantages of using causal representation. On one hand, by decomposing the latent variables into unrelated, separate sets $\boldsymbol{\theta}_i$, each of which corresponds to a causal module~\cite{Pearl00}, it is easier to interpret the parameters and find their valid regions corresponding to reasonable data. On the other hand, even if we just use the parameter values learned from observed data, we can easily come up with new data by making use of their combinations. For instance, suppose we have $(\boldsymbol{\theta}_1, \boldsymbol{\theta}_2)$ and $(\boldsymbol{\theta}_1', \boldsymbol{\theta}_2')$ learned from two domains. Then $(\boldsymbol{\theta}_1, \boldsymbol{\theta}_2')$ and $(\boldsymbol{\theta}_1', \boldsymbol{\theta}_2)$ also correspond to valid causal processes because of the modularity property of causal process. 

\subsection{Causal Discovery to Determine the CG-DAN Structure}\label{Subsec:CD}
With a single source domain, we adopted a widely-used causal discovery algorithm, PC \cite{Spirtes00}, to learn the causal graph up to the Markov equivalence class. (Note that the target domain does not have $Y$ values, but we aim to find causal structure involving $Y$, so in the causal discovery step we do not use the target domain.) We add the constraint that $Y$ is a root cause for $X_i$ to identify the causal structure on a single source domain. 

If there are multiple source domains, we modify the CD-NOD method \cite{zhang2017causal}, which extends the original PC to the case with distribution shifts. Particularly, the domain index $S$ is added as an additional variable into the causal system to capture the heterogeneity of the underlying causal model. CD-NOD allows us to recover the underlying causal graph robustly and detect changing causal modules. By doing so, we only need to learn $\boldsymbol{\theta}_i$ for the changing modules.

Since the causal discovery methods are not guaranteed to find all edge directions, we propose a simple solution to build a network on top of the produced Partially Directed Acyclic Graph (PDAG), so that we can construct the network structure for CG-DAN. We detect indirectly connected variable groups each of which contains a group of nodes connected by undirected edges, and then form a ``DAG'', with nodes representing individual variables or variables groups. We can apply the network construction procedure described in Sec. \ref{Subsec:net_cons} to this  ``DAG''.


\section{Experiments}\label{main_section_experiemnts}
We evaluate the proposed G-DAN and CG-DAN on both synthetic and real datasets. We test our methods on the rotated MNIST \cite{lecun1998gradient} dataset, the MNIST USPS \cite{denker1989neural} dataset, and the WiFi localization dataset \cite{zhang2013covariate}. The first two image datasets are used to illustrate the ability of G-DAN to generate new sensible domains and perform prediction in the target domain. The WiFi dataset is used to evaluate CG-DAN, demonstrating the advantage of incorporating causal structures in generative modeling. CG-DAN is not applied on image data because images are generated hierarchically from high-level concepts and it is thus not sensible to model the causal relations between pixels.


\paragraph{Implementation Details}
We implement all the models using PyTorch and train the models with the RMSProp optimizer \cite{tieleman2012lecture}. On the two image datasets, we adopt the DCGAN network architecture \cite{radford2015unsupervised}, which is a widely-used convolutional structure to generate images. We use the DCGAN generator for the $g$ function in our G-DAN model. As done in \cite{li2017mmd}, we add a discriminator network $f$ to transform the images into high-level features and compare the distributions of the learned features using MMD. We use a mixture of RBF kernels $k(x,x')=\sum_{q=1}^K k_{\sigma_q}(x,x')$ in MMD. On the image datasets, we fix $K=5$ and $\sigma_q$ to be $\{1,2,4,8,16\}$. On the WiFi dataset, we fix $K=5$ and $\sigma_q$ to be $\{0.25,0.5,1,2,4\}$ times the median of the pairwise distances between all source examples.

\subsection{Simulations: Rotated MNIST}
We first conduct simulation studies on the MNIST dataset to demonstrate how distribution changes can be identified from multiple source domains. MNIST is a handwritten digit dataset including ten classes $0-9$ with $60,000$ training images and $10,000$ test images.
We rotate the images by different angles and construct corresponding domains. We denote a domain with images rotated by angle $\gamma$ as $\mathcal{D}_{\gamma}$. The dimensionality of $\boldsymbol{\theta}$ in G-DAN is set to $1$. 

Since $g$ is generally nonlinear w.r.t. $\boldsymbol{\theta}$, one cannot expect full identification of $\boldsymbol{\theta}$ from only two domains. However, we might be able to identify $\boldsymbol{\theta}$ from two nearby domains, if $g$ is approximately linear w.r.t. $\boldsymbol{\theta}$ between these two domains. To verify this, we conduct experiments on two synthetic datasets. One dataset contains two source domains $\mathcal{D}_{0\degree}$ and $\mathcal{D}_{45\degree}$ and the other has two source domains $\mathcal{D}_{0\degree}$ and $\mathcal{D}_{90\degree}$. We train G-DAN to obtain $\hat{g}$ and $\hat{\boldsymbol{\theta}}$ in each domain. 

To investigate whether the model is able to learn meaningful rotational changes, we sample $\boldsymbol{\theta}$ values uniformly from $[\hat{\theta}_s, \hat{\theta}_t]$ to generate new domains. $\hat{\theta}_s$ and $\hat{\theta}_t$ are the learned domain-specific parameters in the source and target domains, respectively. As shown in Figure \ref{fig:rotate45_90}, on the dataset with source domains $\mathcal{D}_{0\degree}$ and $\mathcal{D}_{45\degree}$, our model successfully generates a new domain in between. However, on the dataset with source domains $\mathcal{D}_{0\degree}$ and $\mathcal{D}_{90\degree}$, although our model well fits the two source domains, the generated new domain does not correspond to the domain with rotated images, indicating that the two domains are too different for G-DAN to identify meaningful distribution changes. 

\begin{figure}[ht!]
  \setlength{\abovecaptionskip}{0pt}
\setlength{\belowcaptionskip}{0pt}
\vspace{-3mm}
\begin{center}

\begin{subfigure}{0.3\textwidth}
  \begin{center}
  \includegraphics[height=3.5cm, width=3.5cm]{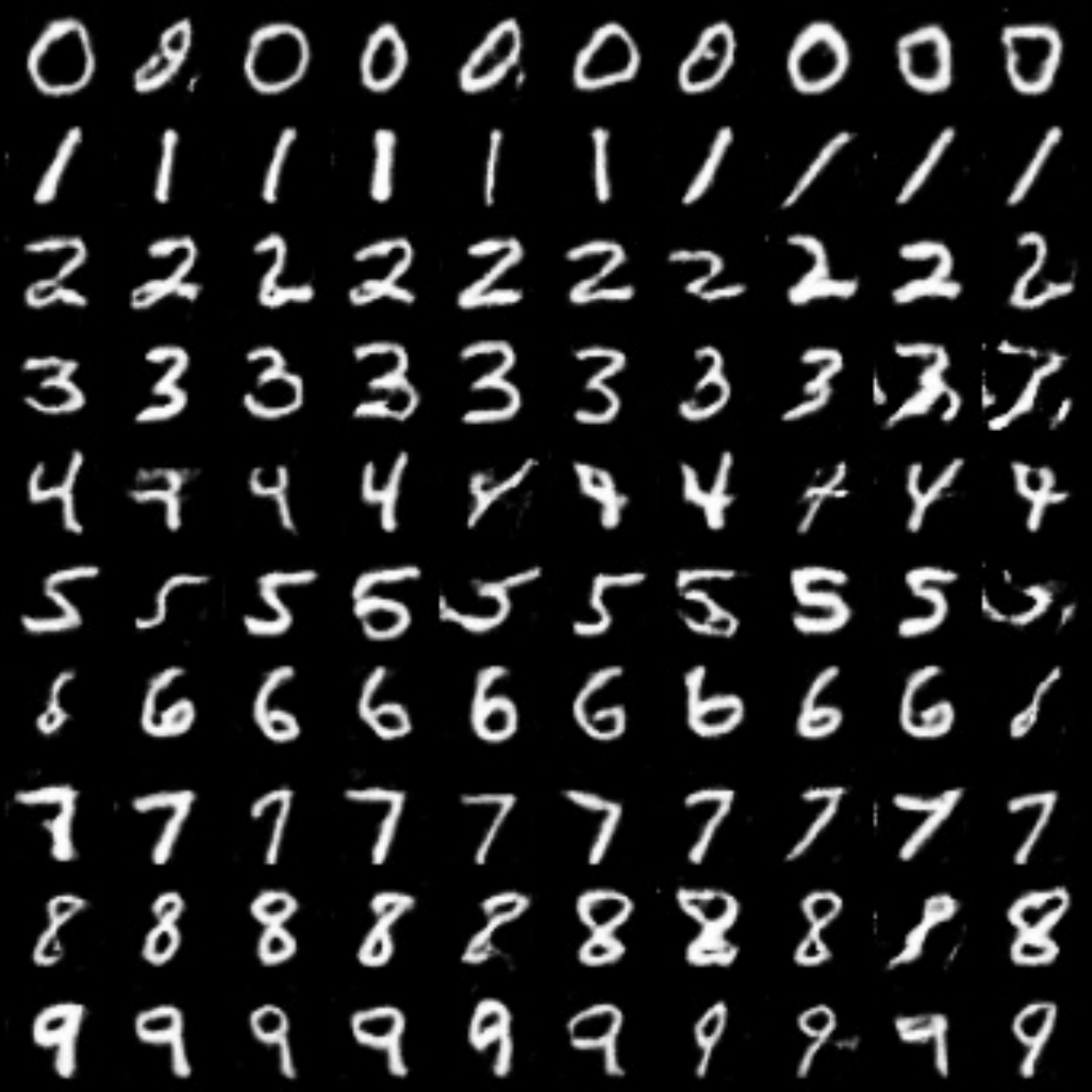}
  \caption*{\footnotesize{Source($0\degree$)}}
  \end{center}
\end{subfigure}%
\hspace{0cm}
\begin{subfigure}{0.3\textwidth}
  \begin{center}
  \includegraphics[height=3.5cm, width=3.5cm]{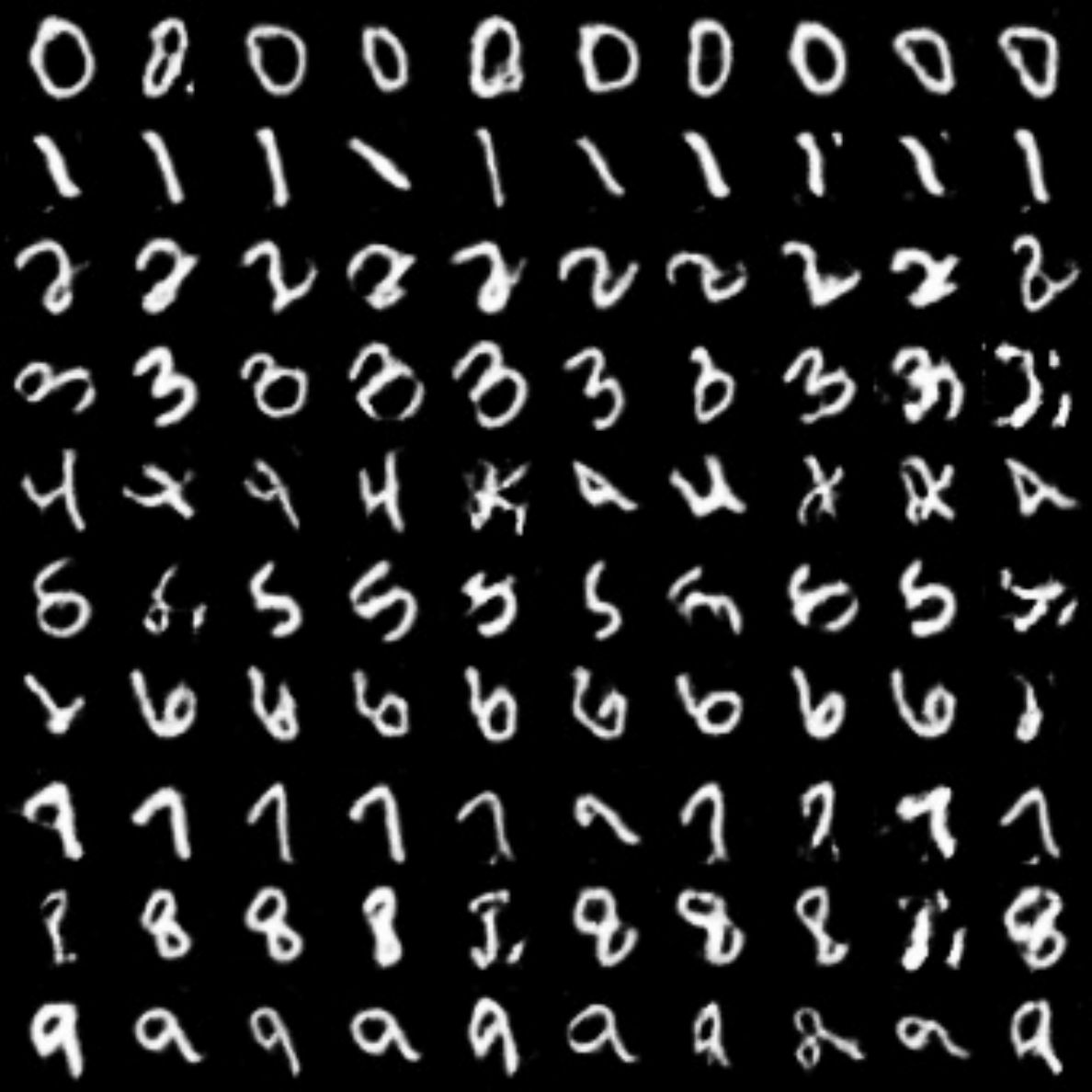}
  \caption*{\footnotesize{New Domain}}
  \end{center}
\end{subfigure}%
\hspace{0cm}
\begin{subfigure}{0.3\textwidth}
  \begin{center}
  \includegraphics[height=3.5cm, width=3.5cm]{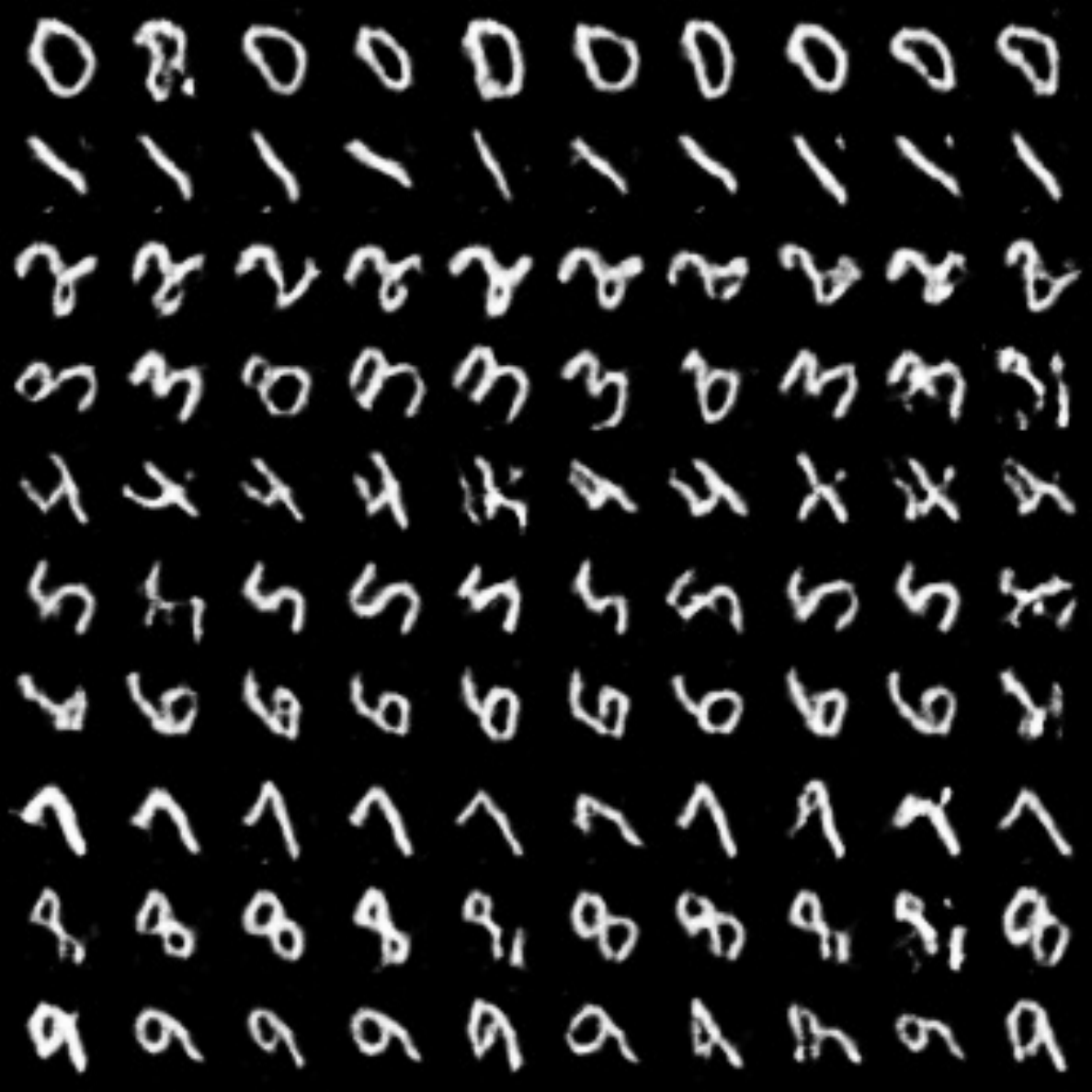}
  \caption*{\footnotesize{Source($45\degree$)}}
  \end{center}
\end{subfigure}%
\hspace{0cm}

\begin{subfigure}{0.3\textwidth}
  \begin{center}
  \includegraphics[height=3.5cm, width=3.5cm]{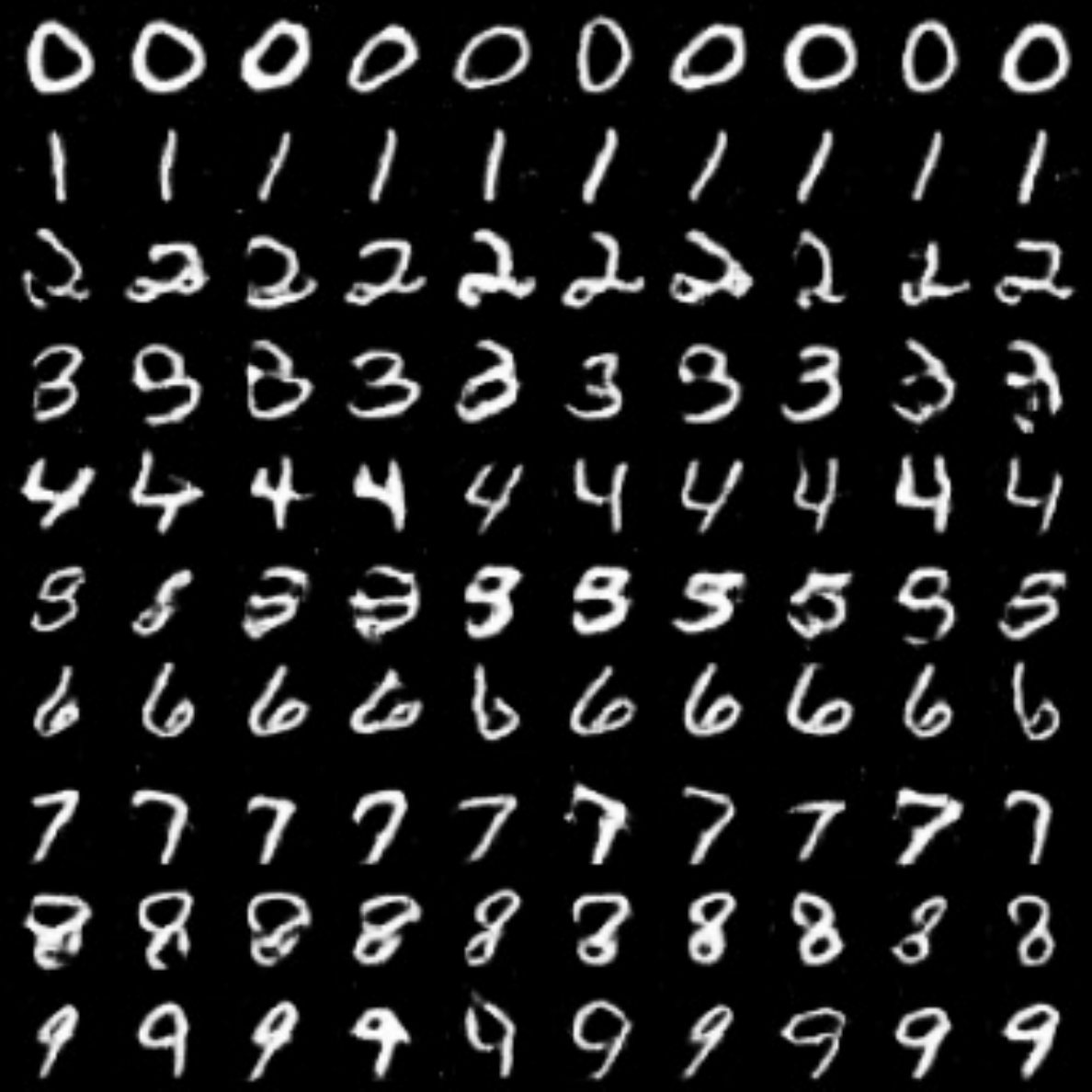}
  \caption*{\footnotesize{Source($0\degree$)}}
  \end{center}
\end{subfigure}%
\hspace{0cm}
\begin{subfigure}{0.3\textwidth}
  \begin{center}
  \includegraphics[height=3.5cm, width=3.5cm]{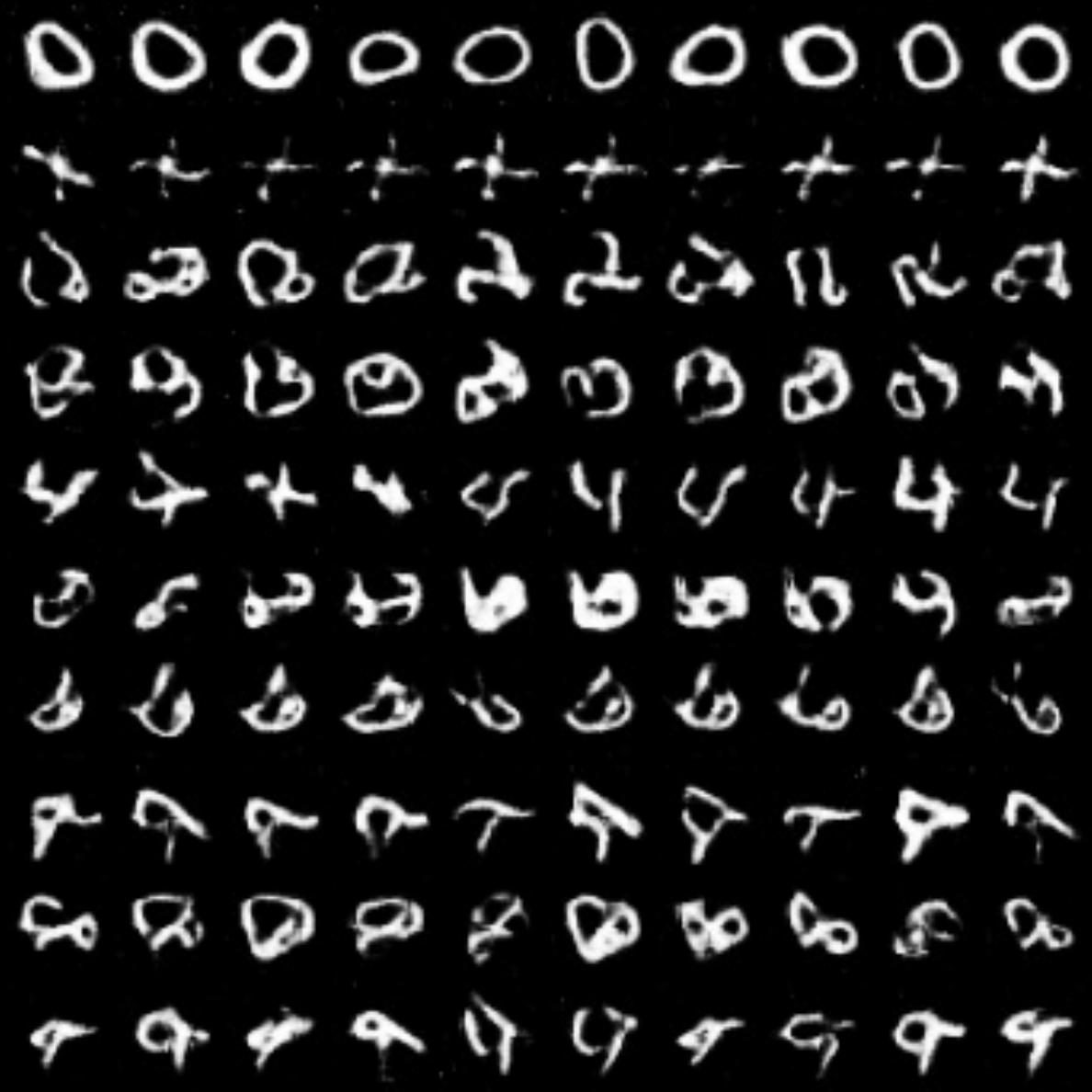}
  \caption*{\footnotesize{New Domain}}
  \end{center}
\end{subfigure}%
\hspace{0cm}
\begin{subfigure}{0.3\textwidth}
  \begin{center}
  \includegraphics[height=3.5cm, width=3.5cm]{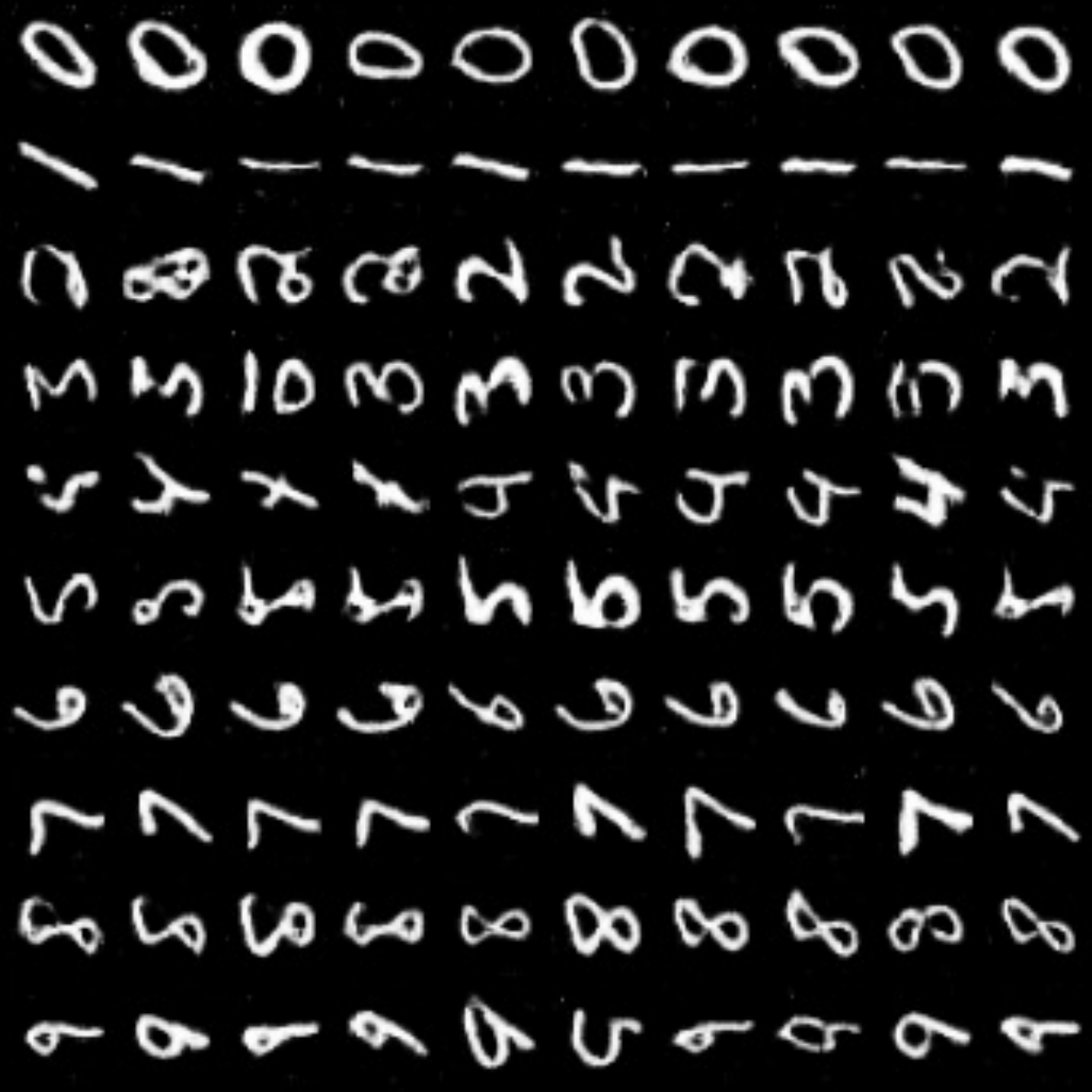}
  \caption*{\footnotesize{Source($90\degree$})}
  \end{center}

\end{subfigure}%

\caption{\small{Generated images from our model. The first row shows the generated images in source domains $\mathcal{D}_{0\degree}$ and $\mathcal{D}_{45\degree}$ and the new domain. The second row shows the generated images in source domains $\mathcal{D}_{0\degree}$ and $\mathcal{D}_{90\degree}$ and the new domain. An animated illustration is provided in Supplementary Material.}}
\label{fig:rotate45_90}
\vspace{-0.5cm}
\end{center}
\end{figure}

\subsection{MNIST-USPS}

Here we use MNIST-USPS dataset to demonstrate whether G-DAN can successfully learn distribution changes and generate new domains. We also test the classification accuracy in the target domain. USPS is another handwritten digit dataset including ten classes $0-9$ with $7,291$ training images and $2,007$ test images. There exists a slight scale change between the two domains. Following CoGAN \cite{liu2016coupled,bousmalis2016domain}, we use the standard training-test splits for both MNIST and USPS. We compare our method with CORAL \cite{sun2016deep}, DAN \cite{icml2015_long15}, DANN \cite{ganin2016domain}, DSN \cite{bousmalis2016domain}, and CoGAN \cite{liu2016coupled}. We adopt the discriminator in CoGAN for classification by training on the generated labeled images from our G-DAN model. The quantitative results are shown in Table \ref{Table:mnist2usps}. It can be seen that our method achieves slightly better performance than CoGAN and outperforms the other methods. 

\begin{figure}[ht!]

\begin{center}

\begin{subfigure}{0.3\textwidth}
  \begin{center}
  \includegraphics[height=3.5cm, width=3.5cm]{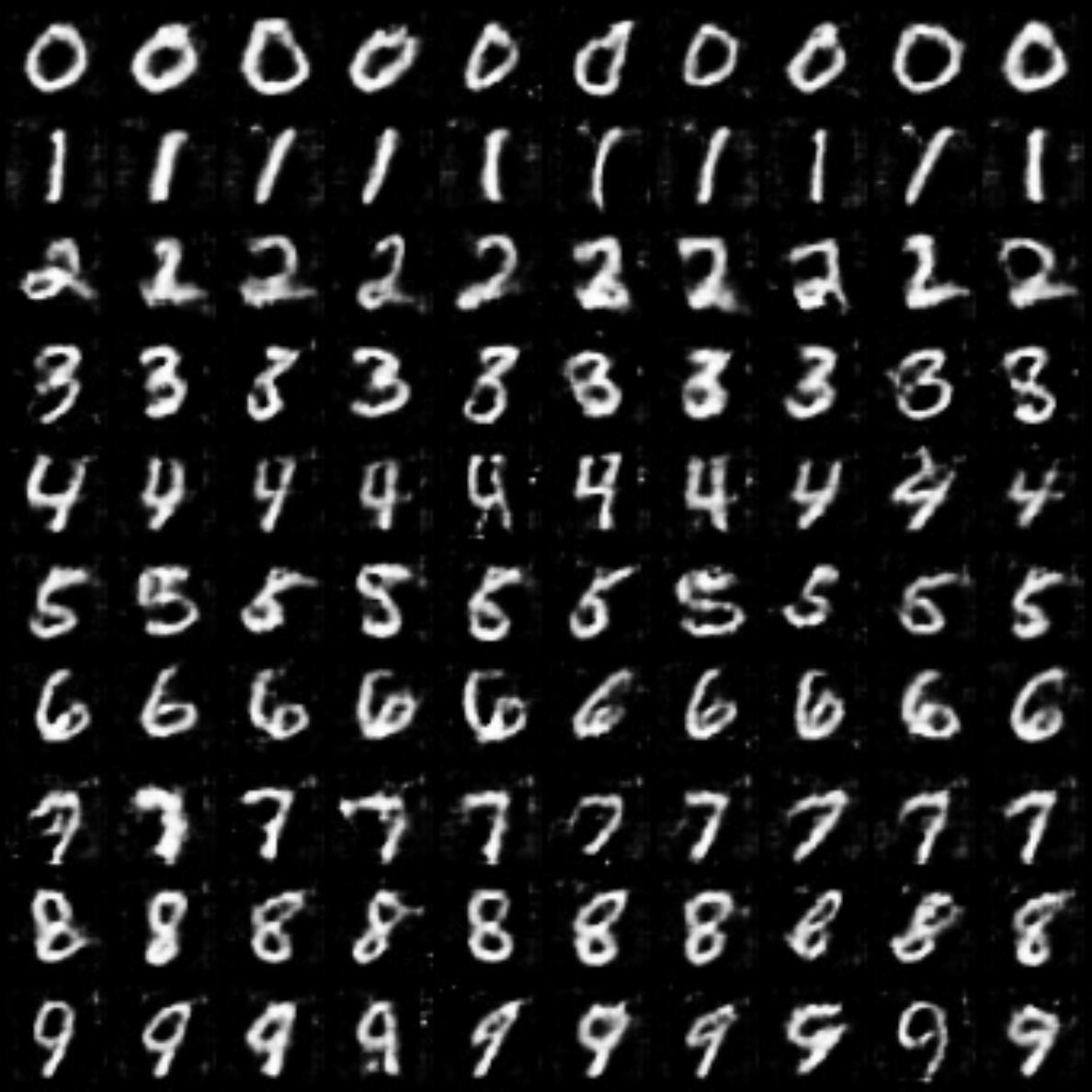}
  \caption*{\footnotesize{Source (MNIST)}}
  \end{center}
\end{subfigure}%
\hspace{0cm}
\begin{subfigure}{0.3\textwidth}
  \begin{center}
  \includegraphics[height=3.5cm, width=3.5cm]{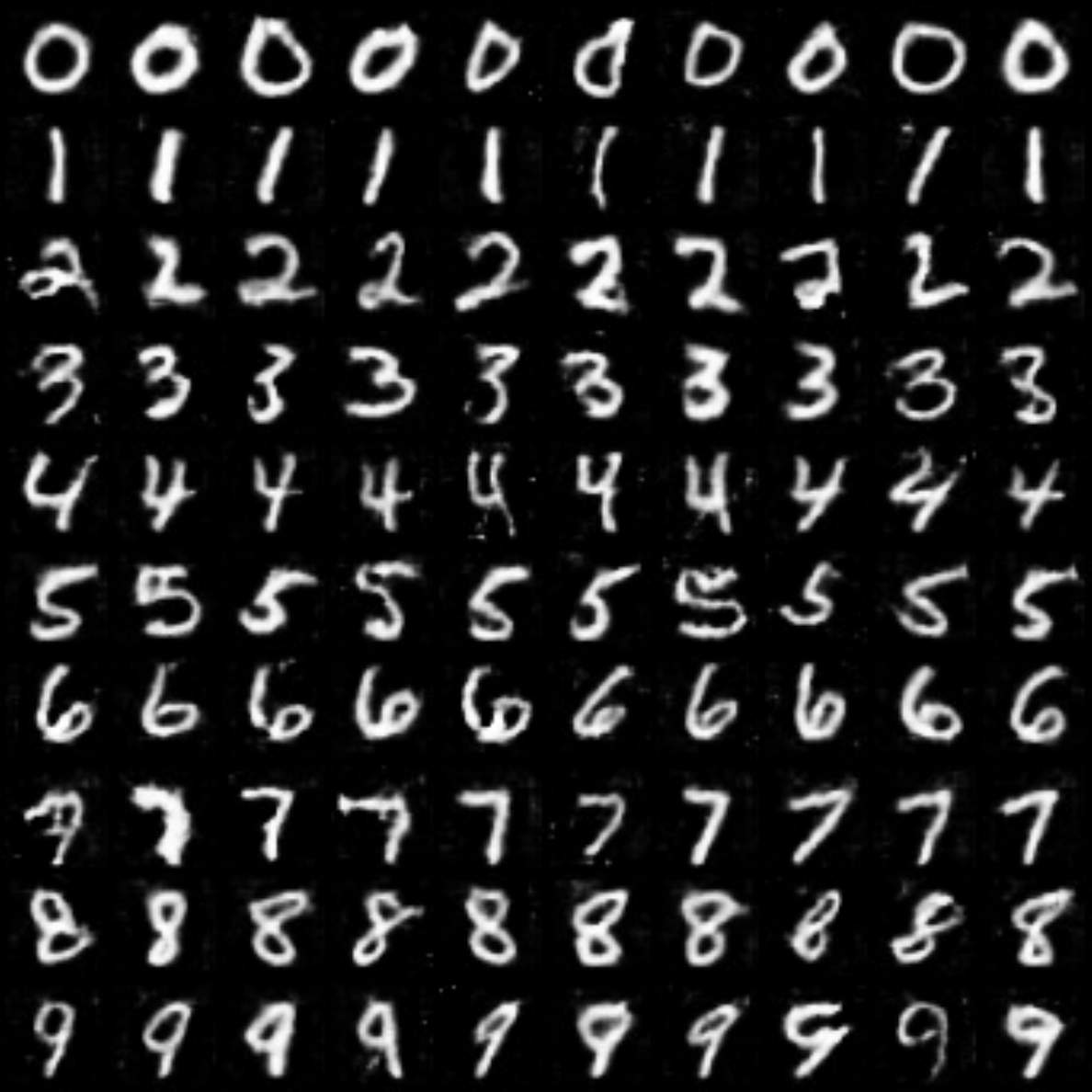}
  \caption*{\footnotesize{New}}
  \end{center}
\end{subfigure}%
\hspace{0cm}
\begin{subfigure}{0.3\textwidth}
  \begin{center}
  \includegraphics[height=3.5cm, width=3.5cm]{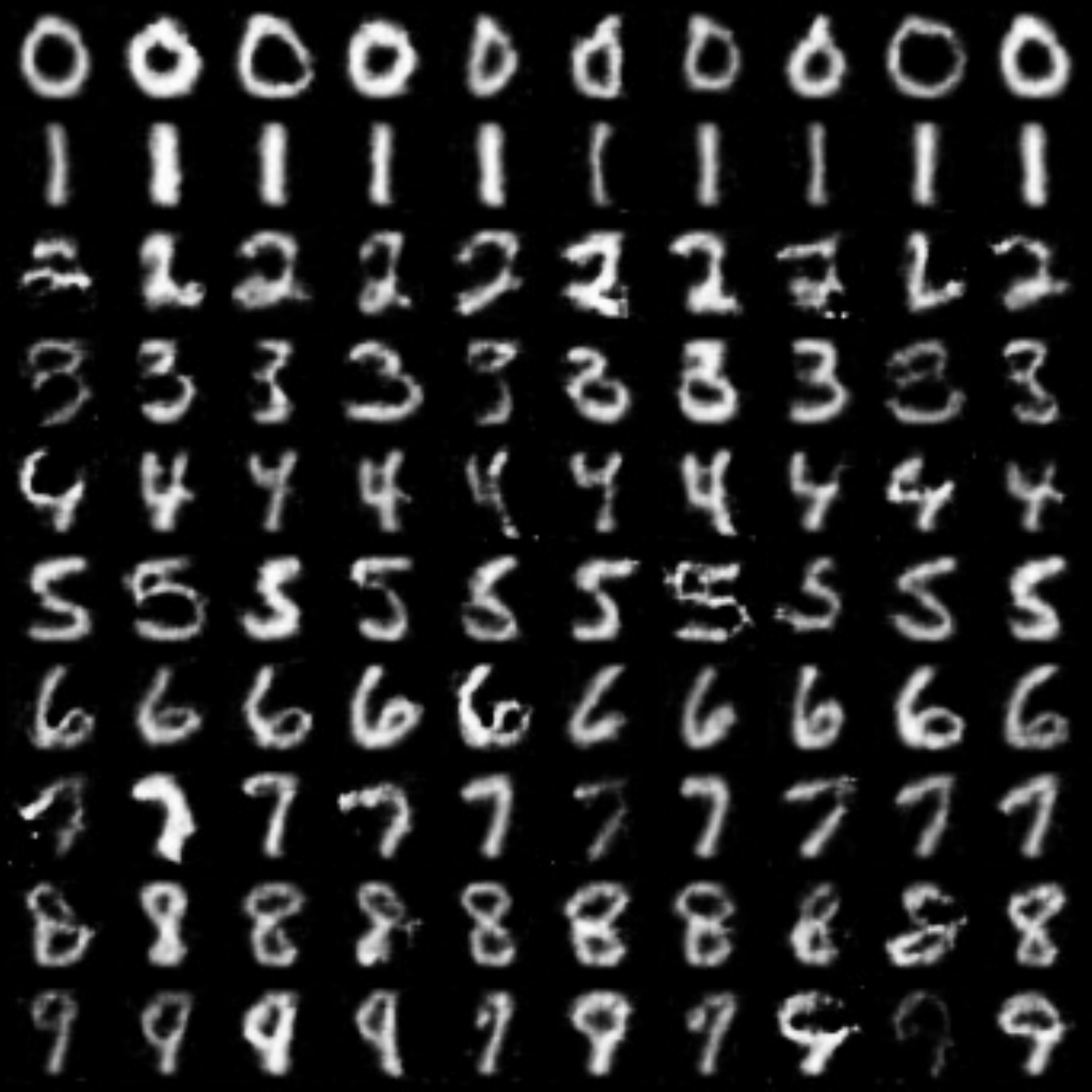}
  \caption*{\footnotesize{Target (USPS)}}
  \end{center}
\end{subfigure}%
 \caption{\small{Generated images on the source, new, and target domain. An animated illustration is provided in Supplementary Material.}}
\label{fig:mnist_usps}
\end{center}
\vspace{-0.4cm}
\end{figure}

In addition, we provide qualitative results to demonstrate our model's ability to generate new domains. As shown in Figure \ref{fig:mnist_usps}, we generate a new domain in the middle of MNIST and USPS. The images on the new domain have slightly larger scale than those on MNIST and slightly smaller scale than those on USPS, indicating that our model understands how the distribution changes. Although the slight scale change is not easily distinguishable by human eyes, it causes a performance degradation in terms of classification accuracy. The proposed G-DAN successfully recovers the joint distribution in the target domain and enables accurate prediction in the target domain. 

\begin{table}[t]
\centering \caption{Comparison of different methods on MNIST-USPS.}
\renewcommand{\arraystretch}{0.4}
{\small
{\begin{tabular}[t]{ccccccc}
\toprule
  CORAL & DAN & DANN & DSN & CoGAN & GDAN\\
 \midrule
  81.7 & 81.1 & 91.3 & 91.2 & 95.7  & \textbf{95.9}\\
\bottomrule
\end{tabular}}}
\label{Table:mnist2usps}
\vspace{-0.1cm}
\end{table}
\subsection{Cross-Domain Indoor WiFi Localization}

We then perform evaluations on the cross-domain indoor WiFi location dataset~\cite{zhang2013covariate} to demonstrate the advantage of incorporating causal structures. The WiFi data were collected from a building hallway area, which was discretized into a space of $119$ grids. At each grid point, the strength of WiFi signals received from $D$ access points was collected. We aim to predict the location of the device from the $D$-dimensional WiFi signals, which casts as a regression problem. The dataset contains two domain adaptation tasks: 1) transfer across time periods and 2) transfer across devices. In the first task, the WiFi data were collected by the same device during three different time periods $\mathtt{t1}$, $\mathtt{t2}$, and $\mathtt{t3}$ in the same hallway. Three subtasks including $\mathtt{t1}\rightarrow \mathtt{t2}$, $\mathtt{t1}\rightarrow \mathtt{t3}$, and $\mathtt{t2}\rightarrow \mathtt{t3}$ are taken for performance evaluation. In the second task, the WiFi data were collected by different devices, causing a distribution shift of the received signals. We evaluate the methods on three datasets, i.e., $\mathtt{hallway1}$, $\mathtt{hallway2}$, $\mathtt{hallway3}$, each of which contains data collected by two different devices.

For both tasks, we implement our G-DAN by using a MLP with one hidden layer ($128$ nodes) for the generator $g$ and set the dimension of input noise $E$ and $\boldsymbol{\theta}$ to $20$ and $1$, respectively. In the time transfer task, because the causal structure is stable across domains, we also apply the proposed CG-DAN constructed according to the learned causal structure from the source domains. Sec. 4 in Supplementary Material shows a causal graph and the detected changing modules obtained by the CD-NOD method on $\mathtt{t1}$ and $\mathtt{t2}$ datasets. We use a MLP with one hidden layer ($64$ nodes) to model each $g_i$. The dimensions of $E$ and $\boldsymbol{\theta}_i$ are set to $1$ for all the modules. 

We also compare with KMM, surrogate kernels (SuK) \cite{zhang2013covariate}, TCA \cite{Pan11}, DIP \cite{6751205}, and CTC \cite{GonZhaLiuTaoSch16}. We follow the evaluation procedures in \cite{zhang2013covariate}. The performances of different methods are shown in Table \ref{Table:wifi}. The reported accuracy is the percentage of examples on which the predicted
location is within 3 or 6 meters from the true location for time transfer and device transfer tasks, respectively. It can be seen that CG-DAN outperforms G-DAN and previous methods in the time transfer task, demonstrating the benefits of incorporating causal structures in generative modeling for domain transfer.
\begin{table*}[t]
\setlength{\abovecaptionskip}{0pt}
\setlength{\belowcaptionskip}{0pt}
\vspace{-1mm}
\centering \caption{Comparison of different methods on the WiFi dataset. The top two performing methods are marked in bold.}
\resizebox{0.95\textwidth}{!}
{\begin{tabular}[t]{ccccccccc}
\toprule
 & KRR & TCA & SuK & DIP  & CTC & G-DAN & CG-DAN\\
 \midrule
$\mathtt{t1}\rightarrow \mathtt{t2}$ &$80.84\pm1.14$ & $86.85\pm1.1$ & $\mathbf{90.36\pm1.22}$ & $87.98\pm2.33$  & $89.36\pm1.78$&  $86.33\pm2.95$&$\mathbf{91.66\pm1.52}$\\
$\mathtt{t1}\rightarrow \mathtt{t3}$ &$76.44\pm2.66$ & $80.48\pm2.73$ & $\mathbf{94.97\pm1.29}$ & $84.20\pm4.29$  & $94.80\pm0.87$& $83.91\pm3.24$&$\mathbf{93.17\pm1.89}$\\
$\mathtt{t2}\rightarrow \mathtt{t3}$ &$67.12\pm1.28$ & $72.02\pm1.32$ & $85.83\pm1.31$ & $80.58\pm2.10$  & $\mathbf{87.92\pm1.87}$& $82.65\pm1.87$ &$\mathbf{89.01\pm2.38}$\\
\midrule
$\mathtt{hallway1}$ & $60.02\pm2.60$ & $65.93\pm0.86$ & $76.36\pm2.44$ & $77.48\pm2.68$  & $\mathbf{86.98\pm2.02}$& $\mathbf{85.50\pm2.92}$&-\\
$\mathtt{hallway2}$ &$49.38\pm2.30$ & $62.44\pm1.25$ & $64.69\pm0.77$ & $\mathbf{78.54\pm1.66}$  & $\mathbf{87.74\pm1.89}$&${76.14\pm2.45}$&-\\
$\mathtt{hallway3}$ &$48.42\pm1.32$ & $59.18\pm0.56$ & $65.73\pm1.57$ & $75.10\pm3.39$  & $\mathbf{82.02\pm2.34}$& $\mathbf{76.04\pm2.55}$&-\\
\bottomrule
\end{tabular}}
\label{Table:wifi}
\vspace{-0.5cm}
\end{table*}

\vspace{-3mm}
\section{Conclusion}
\vspace{-2mm}
We have shown how generative models formulated in particular ways and the causal graph underlying the class label $Y$ and relevant features $X$ can improve domain adaptation in a flexible, nonparametric way. This illustrates some potential advantages of leveraging both data-generating processes and flexible representations such as neural networks. To this end, we first proposed a generative domain adaptation network which is able to understand distribution changes and generate new domains. The proposed generative model also demonstrates promising performance in single-source domain adaptation. We then showed that by incorporating reasonable causal 
structure into the model and making use of modularity, one can benefit from a reduction of model complexity and accordingly improve the transfer efficiency. In future work we will study the effect of changing class priors across domains and how to quantify the level of ``transferability'' with the proposed methods.

\newpage
\bibliography{term_fin1}

\begin{thebibliography}{10}

\bibitem{pan2010survey}
Sinno~Jialin Pan and Qiang Yang.
\newblock A survey on transfer learning.
\newblock {\em Knowledge and Data Engineering, IEEE Transactions on},
  22(10):1345--1359, 2010.

\bibitem{Jiang08}
J.~Jiang.
\newblock {\em A literature survey on domain adaptation of statistical
  classifiers}, 2008.

\bibitem{Shimodaira00}
H.~Shimodaira.
\newblock Improving predictive inference under covariate shift by weighting the
  log-likelihood function.
\newblock {\em Journal of Statistical Planning and Inference}, 90:227--244,
  2000.

\bibitem{Huang07}
J.~Huang, A.~Smola, A.~Gretton, K.~Borgwardt, and B.~Sch{\"o}lkopf.
\newblock Correcting sample selection bias by unlabeled data.
\newblock In {\em NIPS 19}, pages 601--608, 2007.

\bibitem{Sugiyama08}
M.~Sugiyama, T.~Suzuki, S.~Nakajima, H.~Kashima, P.~von B{\"u}nau, and
  M.~Kawanabe.
\newblock Direct importance estimation for covariate shift adaptation.
\newblock {\em Annals of the Institute of Statistical Mathematics},
  60:699--746, 2008.

\bibitem{cortes2008sample}
Corinna Cortes, Mehryar Mohri, Michael Riley, and Afshin Rostamizadeh.
\newblock Sample selection bias correction theory.
\newblock In {\em International Conference on Algorithmic Learning Theory},
  pages 38--53. Springer, 2008.

\bibitem{Pan11}
S.~J. Pan, I.~W. Tsang, J.~T. Kwok, and Q.~Yang.
\newblock Domain adaptation via transfer component analysis.
\newblock {\em IEEE Transactions on Neural Networks}, 22:199--120, 2011.

\bibitem{6751205}
M.~Baktashmotlagh, M.T. Harandi, B.C. Lovell, and M.~Salzmann.
\newblock Unsupervised domain adaptation by domain invariant projection.
\newblock In {\em Computer Vision (ICCV), 2013 IEEE International Conference
  on}, pages 769--776, Dec 2013.

\bibitem{ganin2016domain}
Yaroslav Ganin, Evgeniya Ustinova, Hana Ajakan, Pascal Germain, Hugo
  Larochelle, Fran{\c{c}}ois Laviolette, Mario Marchand, and Victor Lempitsky.
\newblock Domain-adversarial training of neural networks.
\newblock {\em JMLR}, 17(1):2096--2030, 2016.

\bibitem{Zhang13_targetshift}
K.~Zhang, B.~Sch{\"o}lkopf, K.~Muandet, and Z.~Wang.
\newblock Domain adaptation under target and conditional shift.
\newblock In {\em ICML}, 2013.

\bibitem{GonZhaLiuTaoSch16}
M.~Gong, K.~Zhang, T.~Liu, D.~Tao, C.~Glymour, and B.~Sch{\"o}lkopf.
\newblock Domain adaptation with conditional transferable components.
\newblock In {\em ICML}, volume~48, pages 2839--2848, 2016.

\bibitem{Pearl00}
J.~Pearl.
\newblock {\em Causality: Models, Reasoning, and Inference}.
\newblock Cambridge University Press, Cambridge, 2000.

\bibitem{Cortes10}
C.~Cortes, Y.~Mansour, and M.~Mohri.
\newblock Learning bounds for importance weighting.
\newblock In {\em NIPS 23}, 2010.

\bibitem{yu2012analysis}
Y.~Yu and C.~Szepesv{\'a}ri.
\newblock Analysis of kernel mean matching under covariate shift.
\newblock In {\em ICML}, pages 607--614, 2012.

\bibitem{courty2017joint}
Nicolas Courty, R{\'e}mi Flamary, Amaury Habrard, and Alain Rakotomamonjy.
\newblock Joint distribution optimal transportation for domain adaptation.
\newblock In {\em NIPS}, pages 3733--3742, 2017.

\bibitem{icml2015_long15}
M.~Long, Y.~Cao, J.~Wang, and M.~Jordan.
\newblock Learning transferable features with deep adaptation networks.
\newblock In David Blei and Francis Bach, editors, {\em ICML}, pages 97--105.
  JMLR Workshop and Conference Proceedings, 2015.

\bibitem{Storkey09}
A.~Storkey.
\newblock When training and test sets are different: Characterizing learning
  transfer.
\newblock In J.~Candela, M.~Sugiyama, A.~Schwaighofer, and N.~Lawrence,
  editors, {\em Dataset Shift in Machine Learning}, pages 3--28. MIT Press,
  2009.

\bibitem{Iyer14}
A.~Iyer, A.~Nath, and S.~Sarawagi.
\newblock Maximum mean discrepancy for class ratio estimation: Convergence
  bounds and kernel selection.
\newblock In {\em ICML}, 2014.

\bibitem{long2017deep}
Mingsheng Long, Han Zhu, Jianmin Wang, and Michael~I Jordan.
\newblock Deep transfer learning with joint adaptation networks.
\newblock In {\em International Conference on Machine Learning}, pages
  2208--2217, 2017.

\bibitem{Spirtes00}
P.~Spirtes, C.~Glymour, and R.~Scheines.
\newblock {\em Causation, Prediction, and Search}.
\newblock MIT Press, Cambridge, MA, 2nd edition, 2001.

\bibitem{goudet2017learning}
Olivier Goudet, Diviyan Kalainathan, Philippe Caillou, David Lopez-Paz,
  Isabelle Guyon, Michele Sebag, Aris Tritas, and Paola Tubaro.
\newblock Learning functional causal models with generative neural networks.
\newblock {\em arXiv preprint arXiv:1709.05321}, 2017.

\bibitem{hoel1954introduction}
Paul~G Hoel et~al.
\newblock Introduction to mathematical statistics.
\newblock {\em Introduction to mathematical statistics.}, (2nd Ed), 1954.

\bibitem{goodfellow2014generative}
Ian Goodfellow, Jean Pouget-Abadie, Mehdi Mirza, Bing Xu, David Warde-Farley,
  Sherjil Ozair, Aaron Courville, and Yoshua Bengio.
\newblock Generative adversarial nets.
\newblock In {\em NIPS}, pages 2672--2680, 2014.

\bibitem{li2015generative}
Yujia Li, Kevin Swersky, and Rich Zemel.
\newblock Generative moment matching networks.
\newblock In {\em ICML}, pages 1718--1727, 2015.

\bibitem{Song13_embedding}
L.~Song, K.~Fukumizu, and A.~Gretton.
\newblock Kernel embeddings of conditional distributions.
\newblock {\em IEEE Signal Processing Magazine}, 30:98 -- 111, 2013.

\bibitem{zhang2017causal}
Kun Zhang, Biwei Huang, Jiji Zhang, Clark Glymour, and Bernhard Sch{\"o}lkopf.
\newblock Causal discovery from nonstationary/heterogeneous data: Skeleton
  estimation and orientation determination.
\newblock In {\em IJCAI}, volume 2017, page 1347, 2017.

\bibitem{lecun1998gradient}
Yann LeCun, L{\'e}on Bottou, Yoshua Bengio, and Patrick Haffner.
\newblock Gradient-based learning applied to document recognition.
\newblock {\em Proceedings of the IEEE}, 86(11):2278--2324, 1998.

\bibitem{denker1989neural}
John~S Denker, WR~Gardner, Hans~Peter Graf, Donnie Henderson, RE~Howard,
  W~Hubbard, Lawrence~D Jackel, Henry~S Baird, and Isabelle Guyon.
\newblock Neural network recognizer for hand-written zip code digits.
\newblock In {\em NIPS}, pages 323--331, 1989.

\bibitem{zhang2013covariate}
Zhang. {Kai}, V.~Zheng, Q.~Wang, J.~Kwok, Q.~Yang, and I.~Marsic.
\newblock Covariate shift in hilbert space: A solution via sorrogate kernels.
\newblock In {\em ICML}, pages 388--395, 2013.

\bibitem{tieleman2012lecture}
Tijmen Tieleman and Geoffrey Hinton.
\newblock Lecture 6.5-rmsprop: Divide the gradient by a running average of its
  recent magnitude.
\newblock {\em COURSERA: Neural networks for machine learning}, 4(2):26--31,
  2012.

\bibitem{radford2015unsupervised}
Alec Radford, Luke Metz, and Soumith Chintala.
\newblock Unsupervised representation learning with deep convolutional
  generative adversarial networks.
\newblock {\em arXiv preprint arXiv:1511.06434}, 2015.

\bibitem{li2017mmd}
Chun-Liang Li, Wei-Cheng Chang, Yu~Cheng, Yiming Yang, and Barnab{\'a}s
  P{\'o}czos.
\newblock Mmd gan: Towards deeper understanding of moment matching network.
\newblock {\em arXiv preprint arXiv:1705.08584}, 2017.

\bibitem{liu2016coupled}
Ming-Yu Liu and Oncel Tuzel.
\newblock Coupled generative adversarial networks.
\newblock In {\em NIPS}, pages 469--477, 2016.

\bibitem{bousmalis2016domain}
Konstantinos Bousmalis, George Trigeorgis, Nathan Silberman, Dilip Krishnan,
  and Dumitru Erhan.
\newblock Domain separation networks.
\newblock In {\em NIPS}, pages 343--351, 2016.

\bibitem{sun2016deep}
Baochen Sun and Kate Saenko.
\newblock Deep coral: Correlation alignment for deep domain adaptation.
\newblock In {\em ECCV}, pages 443--450. Springer, 2016.

\end{thebibliography}
\bibliographystyle{unsrt}

\newpage
\begin{center}{\LARGE Supplement to \\``Causal Generative Domain Adaptation Networks"}\end{center}
 \vspace{.6cm} {\large This supplementary material provides the
 proofs and some details which are omitted in the submitted paper. The equation
 numbers in this material are consistent with those in the paper.}\\~~\\
\section*{S1. Proof of Proposition 1}
\begin{proof}
$\hat{{\theta}}_1=\hat{{\theta}}_2$ implies that $Q_{X|Y;\hat{\boldsymbol{\theta}}=\hat{\theta}_1}=Q_{X|Y;\hat{\boldsymbol{\theta}}=\hat{\theta}_2}$. Matching ${P}^s_{X|Y}$ and $Q^s_{X|Y}$ results in $P_{X|Y;\boldsymbol{\eta}}=Q_{X|Y;\hat{\boldsymbol{\theta}}}$. Thus, we have $P_{X|Y;\boldsymbol{\eta}=\eta_1}=P_{X|Y;{\boldsymbol{\eta}}=\eta_2}$. Due to the identifiability of $\boldsymbol{\eta}$ in $P_{X|Y;\boldsymbol{\eta}}$, we further have ${\eta}_1={\eta}_2$.
\end{proof}

\section*{S2. Proof of Proposition 2}
\begin{proof}
 After conditional distribution matching on all domains, we have $Q_{X|Y;\boldsymbol{\theta^*}=\theta^{*^{(s)}},g^*}=Q_{X|Y;\hat{\boldsymbol{\theta}}=\hat{\theta}^{(s)},\hat{g}}$ and thus $E_{Q_{X|Y;\boldsymbol{\theta^*}=\theta^{*^{(s)}},g^*}}[X]=E_{Q_{X|Y;\hat{\boldsymbol{\theta}}=\hat{\theta}^{(s)},\hat{g}}}[X]$ for $s={1,\cdots,m}$. Moreover, because $E_{Q_{X|Y;\boldsymbol{\theta},g}}[X]=A\boldsymbol{\theta}+h(Y)$, for any $y\in\mathcal{Y}$, we have 
 \begin{equation}
 A^*\theta^{s^*}+h^*(y)=\hat{A}\hat{\theta}^{(s)}+\hat{h}(y), ~for ~s={1,\cdots,m},\label{Eq:A_theta}
\end{equation}
which can be written in matrix form:
\begin{equation}
 [A^*, h^*(y)]\begin{bmatrix}\theta^{*^{(1)}},&\ldots,&\theta^{*^{(m)}}\\1,&\cdots,&1\end{bmatrix}=[\hat{A}, \hat{h}(y)]\begin{bmatrix}\hat{\theta}^{(1)},&\ldots,&\hat{\theta}^{(m)}\\1,&\cdots,&1\end{bmatrix}.\label{Eq:A_theta}
\end{equation}
Let $A_{aug}^*=[A^*, h^*(y)]\in \mathbb{R}^{D\times(d+1)}$, $\hat{A}_{aug}=[\hat{A}, \hat{h}(y)]\in \mathbb{R}^{D\times(d+1)}$, $\Theta^*=\begin{bmatrix}\theta^{*^{(1)}},&\ldots,&\theta^{*^{(m)}}\\1,&\cdots,&1\end{bmatrix}$, and $\hat{\Theta}=\begin{bmatrix}\hat{\theta}^{(1)},&\ldots,&\hat{\theta}^{(m)}\\1,&\cdots,&1\end{bmatrix}$. According to the assumptions that $rank(A_{aug}^*)=d+1$ and $rank(\Theta^*)=d+1$, $\hat{A}_{aug}$ and $\hat{\Theta}$ should also have rank $d+1$. Therefore, we have $\boldsymbol{\theta^{*}}=A_{aug}^{*^\dagger}\hat{A}_{aug}\hat{\boldsymbol{\theta}}$ and $\hat{\boldsymbol{\theta}}=\hat{A}_{aug}^{^\dagger}A^*_{aug}\boldsymbol{\theta^{*}}$, indicating that $\boldsymbol{\theta^{*}}$ is a one-to-one mapping of $\hat{\boldsymbol{\theta}}$.

\end{proof}

\section*{S3. Proof of Proposition 3}
\begin{proof}
According to the sum rule, we have 
\begin{flalign}
P_{X|\theta}=\sum_{c=1}^C P_{X|Y=c,\theta}P_{Y=c},\nonumber\\
P_{X|\theta'}=\sum_{c=1}^C P_{X|Y=c,\theta'}P'_{Y=c}.
\end{flalign}
Since $P_{X|\theta}=P_{X|\theta'}$, then $$\sum_{c=1}^C P_{X|Y=c,\theta}P_{Y=c}=\sum_{c=1}^C P_{X|Y=c,\theta'}P'_{Y=c}.$$
Also, because A5 holds true, we have
\begin{equation}\label{Eq:equality_independent}
P_{X|Y=c,\theta}P_{Y=c}-P_{X|Y=c,\theta'}P'_{Y=c}=0.
\end{equation}
Taking the integral of (\ref{Eq:equality_independent}) leads to $P_Y=P'_Y$, which further implies that $P_{X|Y,\theta}=P_{X|Y,\theta'}$.
\end{proof}

\section*{S4. Causal Structure on WiFi Data}
\begin{figure}[ht!]
\begin{center}
  \includegraphics[height=9.5cm, width=12cm]{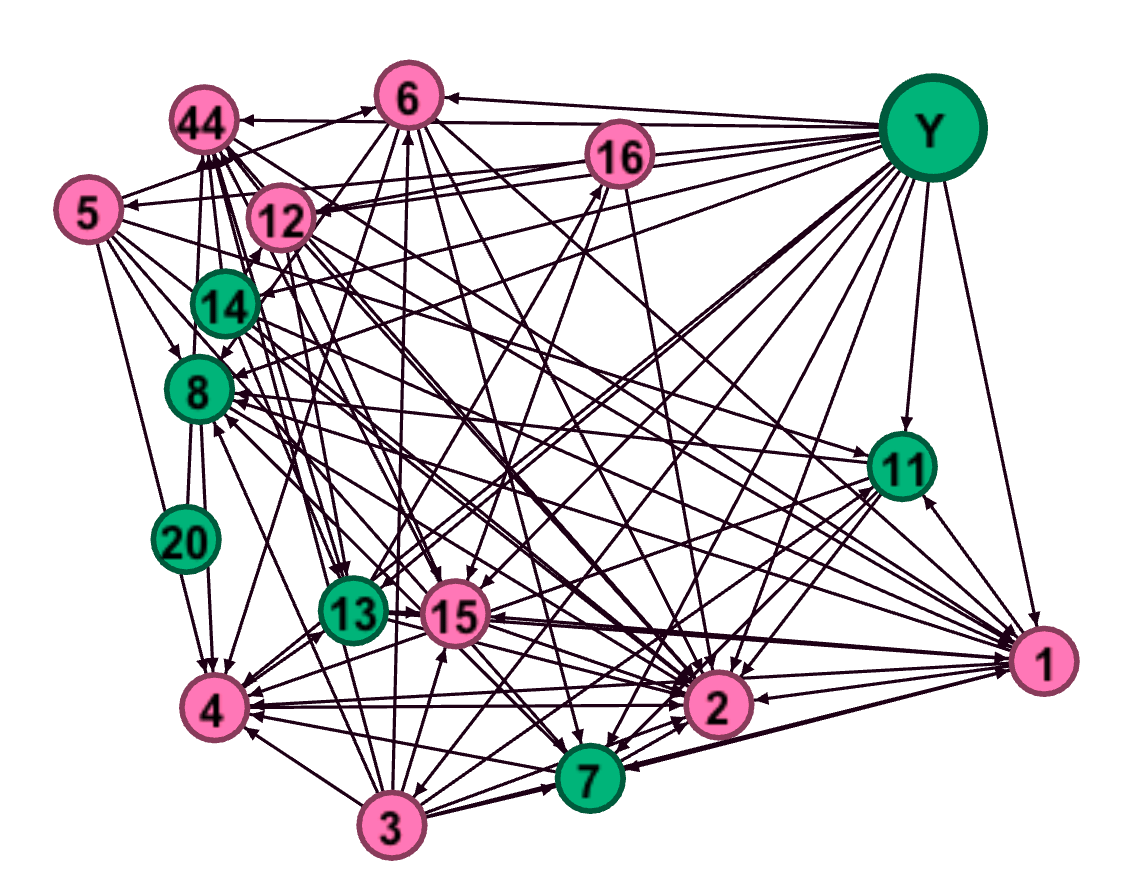}
\caption{\small{The causal structure learned by CD-NOD on the WiFi $\mathtt{t1}$ and $\mathtt{t2}$ datasets. Pink nodes denote the changing modules and green ones denote the constant modules whose conditional distribution does not change across domains.}}
\label{fig:rotate45_90}
\end{center}
\end{figure}
\end{document}